\documentclass{article}

\usepackage{microtype}
\usepackage{graphicx}
\usepackage{subfigure}
\usepackage{booktabs} 
\usepackage{todonotes}
\usepackage{amsmath}

\usepackage{hyperref}

\usepackage[accepted]{icml2019}

\icmltitlerunning{Analyzing Utility of Visual Context in Multimodal Speech Recognition Under Noisy Conditions}

\begin{document}

\twocolumn[
\icmltitle{Analyzing Utility of Visual Context in Multimodal Speech Recognition Under Noisy Conditions}




\icmlsetsymbol{equal}{*}

\begin{icmlauthorlist}
\icmlauthor{Tejas Srinivasan}{lti}
\icmlauthor{Ramon Sanabria}{lti}
\icmlauthor{Florian Metze}{lti}
\end{icmlauthorlist}

\icmlaffiliation{lti}{Language Technologies Institute, Carnegie Mellon University}

\icmlcorrespondingauthor{Tejas Srinivasan}{tsriniva@cs.cmu.edu}

\icmlkeywords{Machine Learning, ICML}

\vskip 0.3in
]




\begin{abstract}
Multimodal learning allows us to leverage information from multiple sources (visual, acoustic and text), similar to our experience of the real world.
However, it is currently unclear to what extent auxiliary modalities improve performance over unimodal models, and under what circumstances the auxiliary modalities are useful.
We examine the utility of the auxiliary visual context in Multimodal Automatic Speech Recognition in adversarial settings, where we deprive the models from partial audio signal during inference time.
Our experiments show that while MMASR models show significant gains over traditional speech-to-text architectures (upto 4.2\% WER improvements), they do not incorporate visual information when the audio signal has been corrupted.
This shows that current methods of integrating the visual modality do not improve model robustness to noise, and we need better visually grounded adaptation techniques.
\end{abstract}

\section{Introduction}


Automatic Speech Recognition (ASR) has been traditionally designed as a unimodal task (\textit{i.e.} acoustic signal as input and text as target). However, our experience of the real world is multimodal - many times, we use the auxiliary modalities to understand the context of a conversation. Based on this idea, previous work has used visual context to adapt ASR models in different ways - we henceforth refer to this as Multimodal ASR (MMASR). More concretely, previous approaches for MMASR systems have centered around individually adapting the acoustic model \cite{miao2016open, moriya2019multimodal} and the language model \cite{gupta2017visual, moriya2018lstm, moriya2019multimodal}. In the acoustic model adaptation, the motivation is that acoustic conditions (\textit{e.g.} car noises, indoor and outdoor acoustics) can be inferred from the scene where the conversation is taking place. In language model adaption, we can use the objects present in the scene to bias the model outputs towards a targeted semantic domain. 

More recently, with the advent of end-to-end ASR models, much work has been centered around incorporating visual context into sequence-to-sequence models \cite{sanabria2018how2, caglayan2019multimodal, palaskar2018end}. While all previous approaches on Multimodal ASR report extensive gains in Word Error Rate (WER) and Perplexity, it is not yet understood where these gains are coming from. This has been observed not only in MMASR, but also in other applications of Multimodal learning such as Neural Machine Translation.

\begin{figure}
    \centering
    \includegraphics[width=0.95\columnwidth]{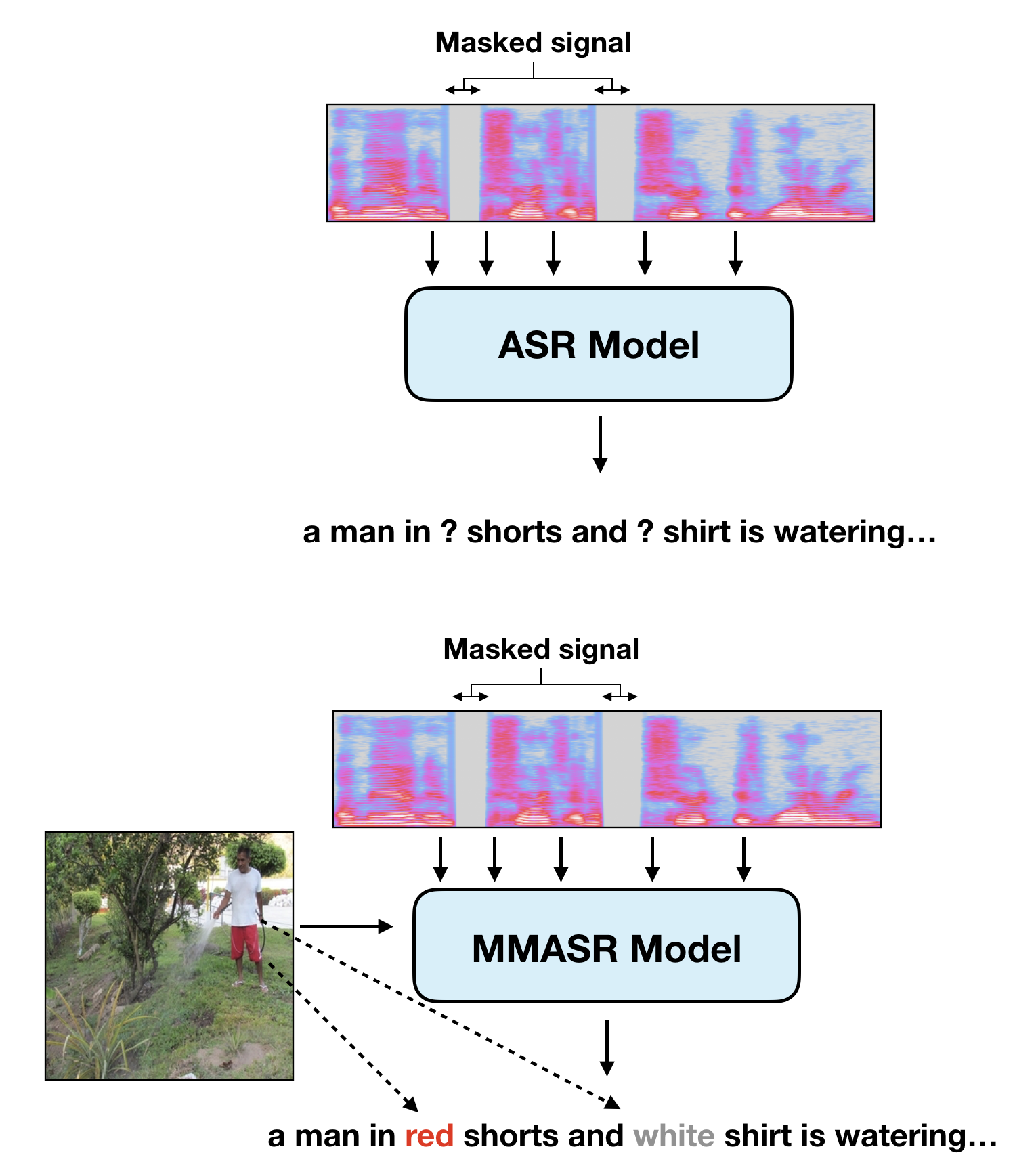}
    \caption{The setup of our MMASR masking experiments.}
    \label{fig:my_label}
\end{figure}

While much of work in Multimodal Machine Translation (MMT) has suggested that the visual modality is marginally beneficial \cite{barrault2018findings, elliott2018adversarial}, recent work \citep{caglayan2019probing} suggests that visual information is useful when there is missing information in the source-side signal. We hypothesize that the same could hold for Multimodal ASR, under conditions when the acoustic speech is corrupted. These corruptions or distortions can be in the form of excessive background noise, or silence during certain segments of the speech utterance. These types of signal anomalies are usually present in real-world ASR applications. We believe that, in such situations, MMASR architectures will use the information from the visual domain to recognize the audio-masked entities. Such behaviour is close to the human psicoacustic behaviour on acoustic perturbation. For instance, \citet{mcgurk1976hearing} suggest that when audio signal is perturbed, humans focus on the visual modality.

\begin{figure*}[t]
    \centering
    \includegraphics[width=0.7\textwidth]{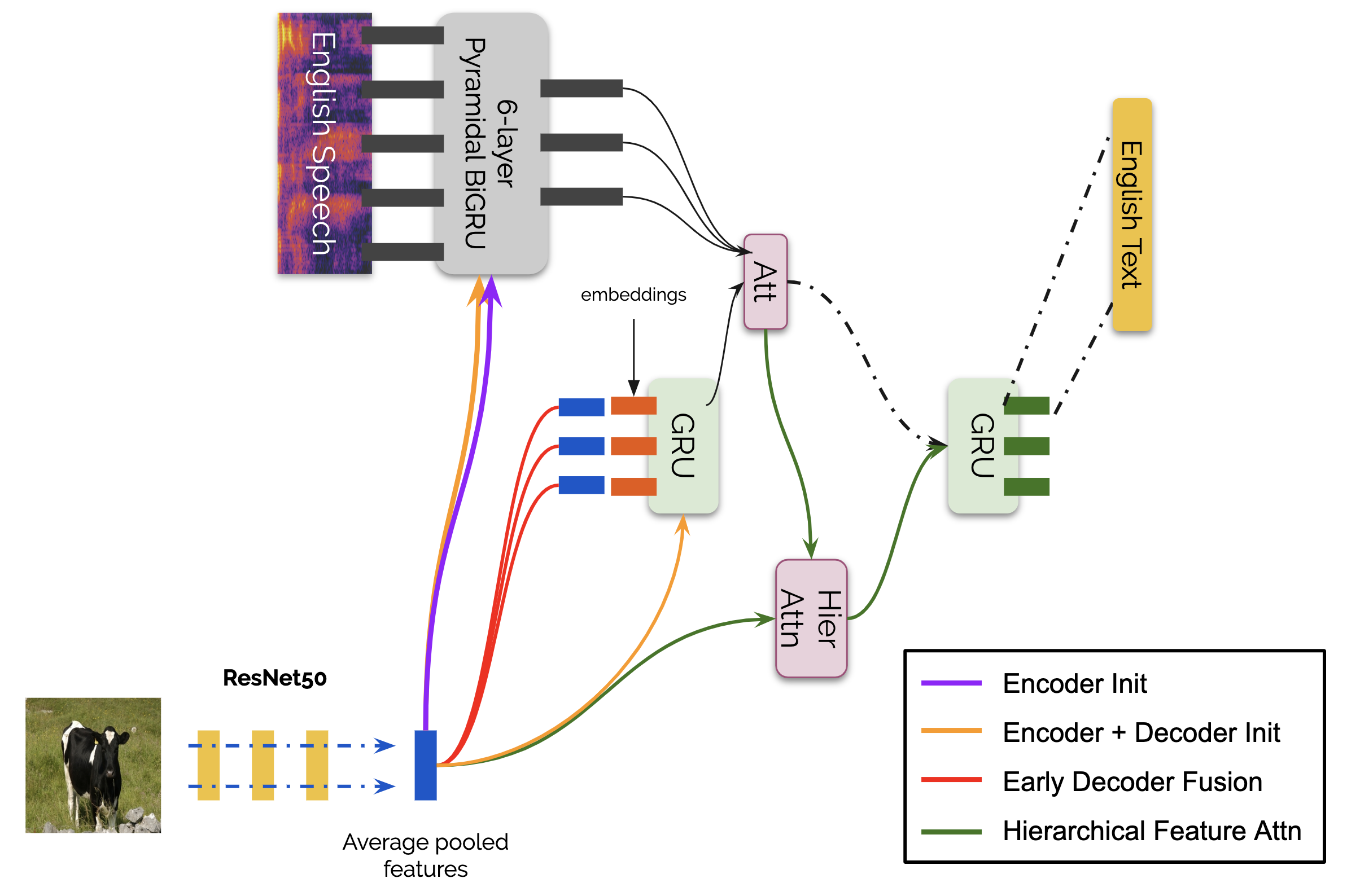}
    \caption{The different variants of our Multimodal ASR models. The colored arrows represent how the visual features are incorporated into the baseline ASR framework in different ways.}
    \label{fig:model}
\end{figure*}


Inspired by \citet{caglayan2019probing}, in this work we port a similar set of experiments to MMASR, where we analyze the contribution of the visual modality to different input signal corruption in the primary modality (\textit{i.e.} acoustic signal) on state-of-the-art MMASR architectures~\cite{sanabria2018how2,caglayan2019multimodal}. Similar to \citet{caglayan2019probing}, we perform three types of masking, by replacing specific words in the acoustic signal with silence during inference time (Section \ref{masking}). We also analyze the sensitivity of the model to the visual modality similar to \citet{elliott2018adversarial}, by deliberately misaligning the audio and visual inputs in our test set (Section \ref{incongruent}). 

Our results (Section \ref{sec:experiments}) on the PlacesAudio dataset \cite{harwath2016unsupervised} show that previously proposed MMASR models, in general, show considerable improvements over the standard unimodal architecture, but fail to incorporate visual information when the audio signal has been corrupted. Moreover, visually-adapted models tend to be less sensitive to image information - even when an unrelated image is fed to the audio input, the MMASR models perform better than the baseline ASR models (albeit slightly worse than correctly-aligned inputs).

\section{Methods}

In this section, we describe the different models we used for Multimodal ASR (Section \ref{models}), and the different masking experiments we performed on the testing data during inference (Section \ref{masking}).

\subsection{Models}
\label{models}
We experiment with a baseline unimodal ASR model, and four multimodal ASR model variants. We can see the baseline ASR model and our different visual adaptation techniques in Figure \ref{fig:model}.

\subsubsection{Baseline ASR}
Our baseline ASR model is a sequence-to-sequence model with attention \cite{bahdanau2014neural, chan2016listen}. Our encoder comprises of 6 bidirectional LSTM layers, each of which is followed by a \textit{tanh} activation. The decoder is a two-layer stacked GRU. A feed-forward attention mechanism over the encoder states $E$ is used after the GRU1 layer to compute the context vector, which is fed as input to the GRU2 layer.
\begin{align*}
    h_t^{dec1} & = GRU_1(y_{t-1}, h_{t-1}^{dec1}) \\
    z_t & = Attention(E, h_t^{1}) \\
    h_t^{dec2} & = GRU_2(z_t, h_t^{dec1}) 
\end{align*}

The probability distribution over the vocabulary is computed through a non-linear transformation of $h_t^2$, followed by a softmax.

\subsubsection{Encoder Initialization}
 The hidden state of the speech encoder is initialized using a non-linear transform of the visual features, $f$.
 \begin{align*}
     h_0^{enc} = tanh(W_v f + b_v)
 \end{align*}

\subsubsection{Encoder + Decoder Initialization}

This is similar to the encoder initialization described above, however, here the first hidden state of the decoder is also initialized by projecting the visual feature vector $f$ to the hidden state dimension, and passing it through an activation function. 
 \begin{align*}
     h_0^{enc} & = tanh(W_e f + b_e) \\
     h_0^{dec1} & = tanh(W_d f + b_d)
 \end{align*}
 
Unlike \citet{caglayan2019multimodal}, we do not share the weights in the projection layer, $W_e$ and $W_d$.

\subsubsection{Early Decoder Fusion}

During decoding, we project the visual features $f$ to the hidden state dimension, and then concatenate it to the input embedding vector $y_t$ at each timestep before passing it through the GRU decoder.
\begin{align*}
    f' & = tanh(W_f f + b_f) \\
    y_{t} & = [y_t; f'] \\
\end{align*}

\subsubsection{Hierarchical Feature Attention}

We compute the speech context vector over the encoder states using the attention mechanism as before. We further learn a second  attention layer over the encoder context vector $z_t$ and the visual features $f$ (which are projected into a common space). This hierarchical attention layer outputs a context vector $z_t^{hier}$, which is fed as input to the GRU2 layer. This is similar to \citet{libovicky2017attention}, but for the visual modality we do not compute an image context vector and use our global visual feature vector $f$ instead.
\begin{align*}
    f' & = tanh(W_p f + b_p) \\
    h_t^{dec1} & = GRU1(y_{t-1}, h_{t-1}^{dec1}) \\
    z_t & = Attention(E, h_t^{1}) \\
    z_t^{hier} & = Attention(\{z_t, f'\}, h_t^{dec1}) \\ 
    h_t^{dec2} & = GRU2(z_t^{hier}, h_t^{dec1}) 
\end{align*}

where $W_p, b_p$ project the visual feature vector to the hidden state dimension. 
\subsection{Masking Experiments}
\label{masking}

We present a number of masking experiments, wherein we replace segments of the audio signal pertaining to specific words with silence.

\subsubsection{Color Masking}

We mask the speech segment corresponding to \textbf{color} words with silence. A total of 21 colors were masked in such a manner in the testing data. This masking affects 2.88\% of words in the testing data, with an average of 0.55 words per test sentence. We expect that the multimodal models should be able to leverage color information from the visual context when the audio signal is corrupted. The color-masked test set is referred to as $T_C$.

\subsubsection{Noun Masking}

We also \textbf{mask select nouns} in each test utterance. We performed Part-of-Speech tagging using the Stanford POS tagger \cite{toutanova2003feature}, and masked words with the \textit{NN} noun tag with a probability of 0.3. This resulted in the masking of 6.27\% of the words in the testing set, with an average of 1.19 words removed per sentence. Since nouns are likely to be seen in the corresponding image, we believe that the auxiliary modality should provide additional context when the speech signal is corrupted. The noun-masked test set is referred to as $T_N$.

\subsubsection{Progressive Masking}

Finally, similar to \citet{caglayan2019probing}, we perform progressive masking by \textbf{progressively removing words from the end of the utterance}. We perform 5 different experiments, where we mask $k$ words from the end of each utterance, where $k \in \{2, 4, 6, 8, 10\}$.

\subsection{Incongruent Decoding}
\label{incongruent}

We test the sensitivity of our multimodal models to visual context, by \textbf{misaligning the auxiliary images} with their corresponding audio input during test time \cite{elliott2018adversarial}. During testing, we randomly select an image from the test set to be the visual context for each test utterance. If a multimodal model is indeed utilizing the visual context, we would expect a noticeable performance drop when the visual modality is unrelated to the audio input.

\section{Experiments}
\label{sec:experiments}
\subsection{Dataset}

We conduct experiments on the PlacesAudio dataset \cite{harwath2016unsupervised}, which consists of more than 400,000 speech utterances for images drawn from the Places205 image dataset. The models are trained on a training split of 402,385 utterances, whereas the validation split provided is divided into equal halves for validation and testing (500 utterances each). Utterances are free-form spoken descriptions of each image, so we believe that visual context should ideally be of great help in the ASR task. Captions in the dataset are already lowercased and tokenized, and no further pre-processing is performed on it.

\subsection{Implementation Details}
All models are trained using Adam optimizer \cite{kingma2014adam}, with a learning rate of 0.0004, decay of 0.5 and batch size of 36. The encoder and decoder GRU both have 256 hidden units. The embedding dimension for the decoder is also 256, and the input and output decoder embedding weights are tied \cite{press2016using}. The norm of the gradient is clipped with a threshold of 1 \cite{pascanu2012understanding}. A dropout of 0.4 is applied on the final encoder and decoder outputs. All models are implemented using the nmtpytorch framework \cite{caglayan2017nmtpy}.

\subsubsection{Audio Features}
We use Kaldi to extract 40-dimensional filter
bank features from 16kHz raw speech signal using a time window of 25ms and an overlap of 10ms. 3-dimensional pitch features are further concatenated to form the final feature vectors.

\subsubsection{Visual Features}
We use a ResNet-50 CNN \cite{he2016deep} for extracting the visual features. Prior to feature extraction, we center and standardize the images using ImageNet statistics, then resize the shortest edge to 256 pixels and take a center crop of size 256x256. We extract spacial features from the final convolutional layer and perform global average pooling to extract 2048-dimensional visual features. Since images in the Places205 dataset are similar to the ImageNet dataset on which ResNet-50 is trained, we do not tune the ResNet-50 weights during training.

\subsection{Results}

\begin{table}[h]
\centering
\begin{tabular}{| l | l | l | l |}
\hline
Model    & $T$ & $T_{C}$ & $T_N$ \\
\hline
Baseline ASR      & 25.1     & 27.9  & 33.5      \\
\hline
Encoder Init & 22.1 &   25.1    & \textbf{30.6}     \\
Encoder + Decoder Init & 22.3 & 25.6  & 30.9   \\
Early Decoder Fusion & 22.0 & 25.1 & 30.7 \\
Hierarchical Feature Attention & \textbf{20.9} & \textbf{24.4} & \textbf{30.6} \\
\hline
\end{tabular}
\label{results}
\caption{Word Error Rate (WER, in \%) results on the PlacesAudio dataset. $T$, $T_C$, $T_N$ are the unmasked, color-masked and noun-masked variants of the test set.}
\end{table}

Experimental results in Table \ref{results} show that all of our multimodal models considerably outperform the unimodal baseline model on the full unmasked test set, each by 2.8-4.2\% WER. The best performance is achieved by the Hierarchical Feature Attention model, which achieves a WER of 20.9\% (an improvement of \textbf{+4.2\%} over the baseline). The Encoder Init and Early Decoder Fusion models perform comparably, while Encoder + Decoder Init performs slightly worse.

When we perform audio masking in the test set, however, we see that performance of the multimodal models does not improve as we had hoped. In the case of color masking, the WER improvement falls slightly, with each model's WER gain falling by 0.2-0.7\% compared to the unmasked performance. The best performing model is still Hierarchical Feature Attention, which has a WER of 24.4\% (\textbf{+3.5\%} improvement over baseline). More importantly, on examining the utterances, none of the models can correctly transcribe color words in the masked signal where the baseline ASR fails. This shows that our visual adaptation techniques fail to fill in the color when the audio signal is corrupted.

Similarly, when masking is applied to the nouns, the multimodal WER gains over the baseline fall by upto 1.2\%. Encoder Init performs as well as the Hierarchical Feature Attention model (\textbf{+2.9\%} improvement over baseline), while the other models perform slightly worse. Similar to the color-masking, the multimodal models do not perform better at transcribing missing nouns than the baseline.

\subsubsection{Progressive Masking}

\begin{figure}[h]
    \centering
    \includegraphics[width=\columnwidth]{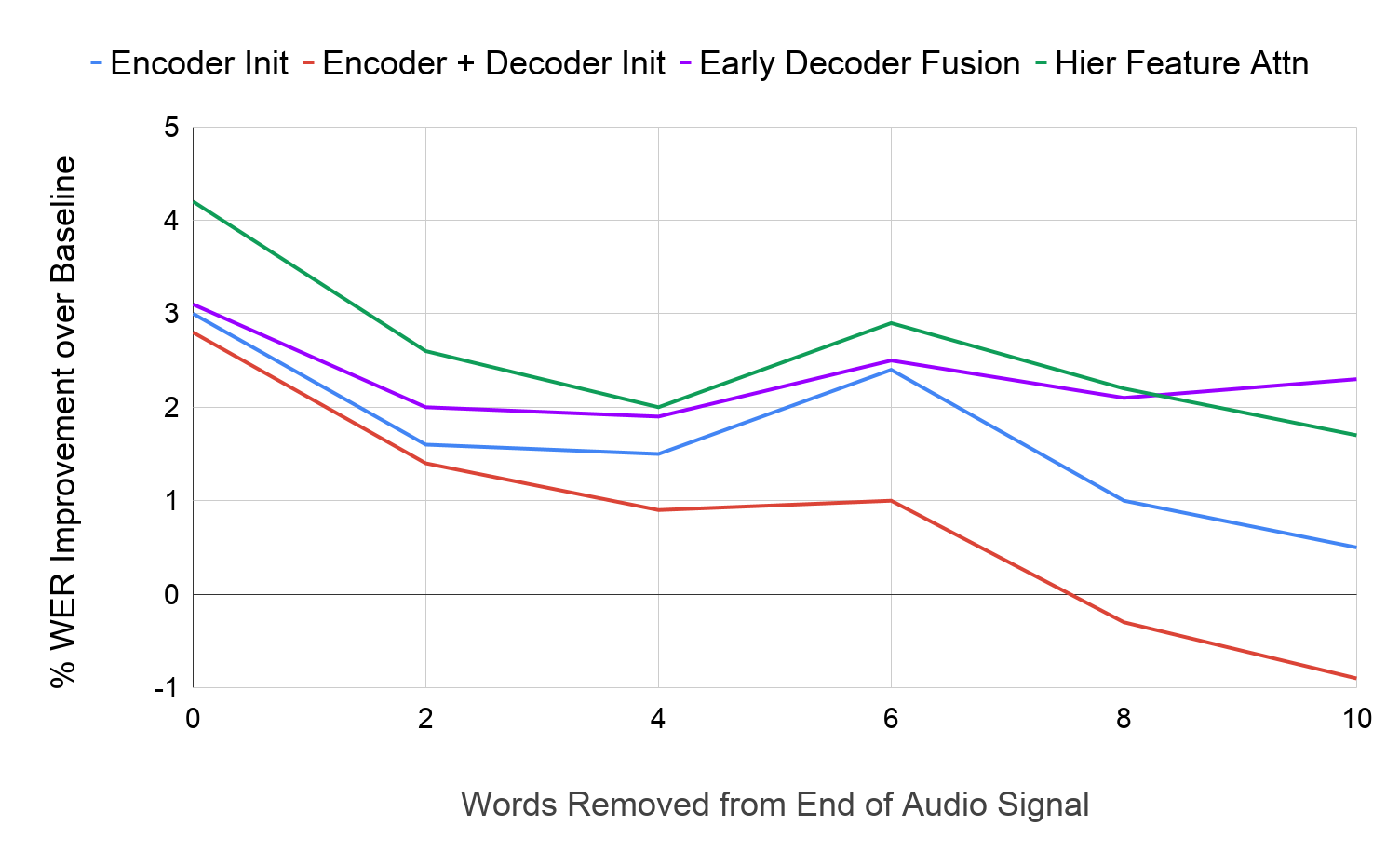}
    \caption{WER improvements of MMASR models when we do progressive masking}
    \label{fig:progressive}
\end{figure}

The results for progressive masking experiments can be seen in Figure \ref{fig:progressive}. As we masking more words from the end of the audio signal, the improvement over baseline generally trends downwards. This is indicative of the fact that multimodal models are not capable of leveraging the visual modality to compensate for the missing signal, and as the amount of masked signal increases, the gains of multimodality goes down.

\subsubsection{Incongruent Decoding}

\begin{figure}[h]
    \centering
    \includegraphics[width=\columnwidth]{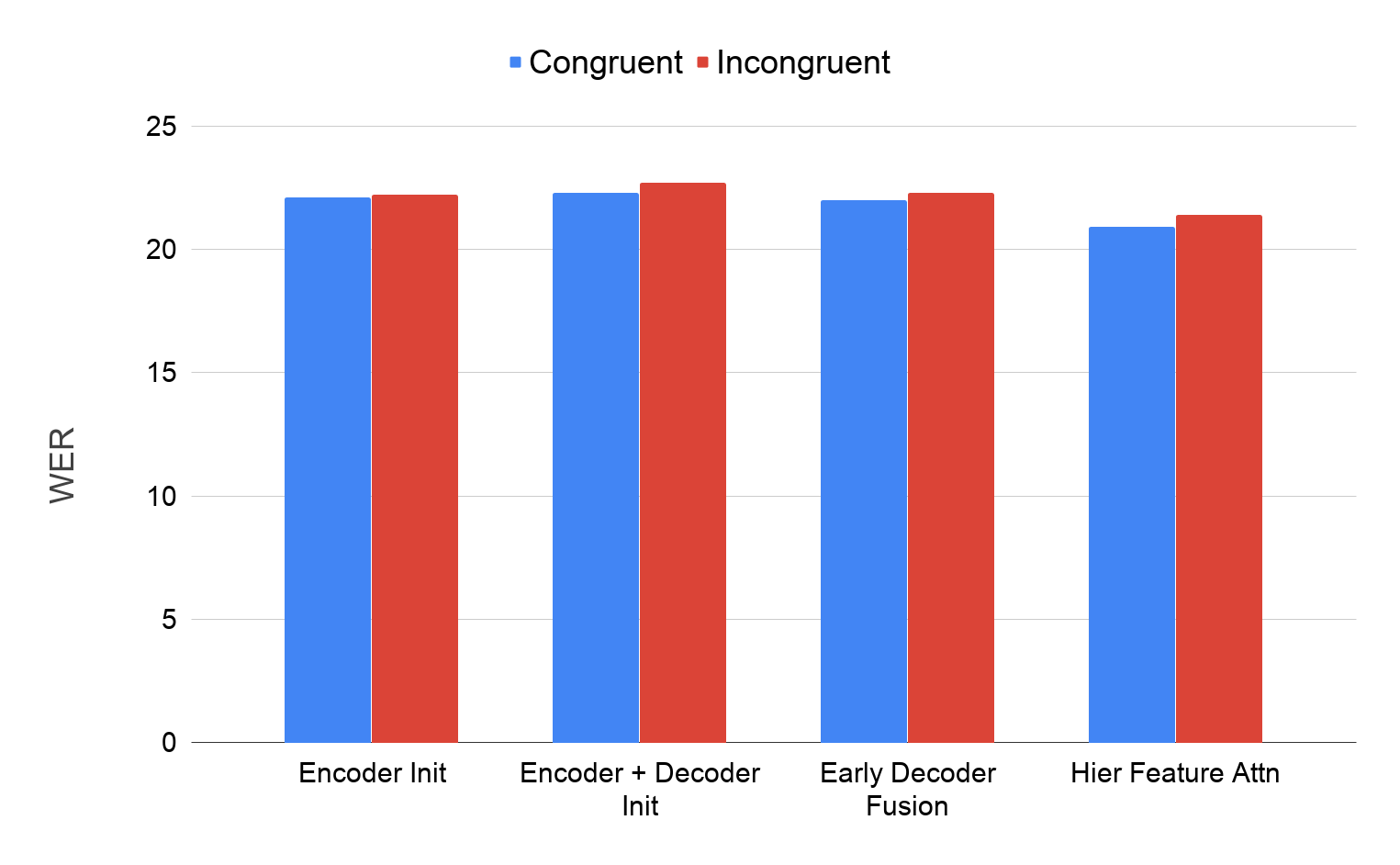}
    \caption{WER of multimodal models in Congruent and Incongruent Decoding scenarios}
    \label{fig:incongruent}
\end{figure}

Finally, we present the results of the Incongruent Decoding experiment in Figure \ref{fig:incongruent}, with all experiments performed on the unmasked test set. For all multimodal models, the WER gets slightly worse when we present an unrelated image to the model during inference (around 0.5\% WER degradation for all models). However, they still perform much better than the baseline ASR models (upto 3.5\% WER improvement). This result in particular tells us that the visual context is not being used correctly, and we need better visual adaptation to understand how the auxiliary modality is being used by the models.

\section{Conclusions and Future Work}
This paper presents several masking experiments to analyze how the visual modality is being utilized in MMASR models. We conclude that while current multimodal ASR models significantly outperform the traditional unimodal ASR model, they are unable to leverage the modality to compensate when the audio signal is corrupted during inference time. More importantly, MMASR models can improve over the unimodal variant even by using an unrelated image as the auxiliary modality, which raises questions about how the visual modality is being used in the first place.

Our findings in this paper open up several new avenues of future research. Current multimodal models are not more robust to noise during inference time, so one line of research would be to introduce similar noise in the training data, and see if multimodal models can learn to leverage the visual modality when they encounter silence. 

Our results also suggest that our global average pooled visual features do not provide sufficient visual context. We also need to experiment with more image-aware multimodal models that can attend over the spatial image features as well as speech features (similar to the hierarchical attention proposed in \citet{libovicky2017attention}). Another solution is to explore more visually grounded models, which are not as end-to-end but encode more explicit information about what is contained in the image.

\section{Related Work}
\label{sec:relatedwork}
Much of the early work in multimodal speech recognition has focused on Audio-Visual Speech Recognition (AVSR), also known as Lip Reading. This is the task of recognizing speech being spoken by a talking face, with or without the audio \cite{chung2017lip, assael2016lipnet}. In the latter case where both audio and visual modalities are present, one of the key challenges has been fusing the two modalities \cite{mroueh2015deep, zhou2019modality}. However, our task differs from AVSR in one key aspect: while the visual domain in lip reading gives context at the phoneme level, images in the PlacesAudio dataset give semantic information about what is being spoken about in the audio input.

Our task is similar to the one described in \citet{sun2016look}, which uses the spoken Flickr8k dataset \cite{harwath2015deep}. In contrast to \citet{sun2016look}, whereas they add an image captioning model to their language model during decoding, we incorporate the visual context directly into our sequence-to-sequence model. Additionally, the PlacesAudio dataset we use is roughly 10 times the size of Flickr8k (403,385 utterances compared to 40,000).

More related to our work on sequence-to-sequence models are \citet{caglayan2019multimodal} and \citet{palaskar2018end}. Both works focus on the task of video subtitling on the How2 dataset \cite{sanabria2018how2}; the visual context in this case varies temporally, unlike in PlacesAudio. \citet{palaskar2018end} does visual adaptation by early fusion, where the visual features are concatenated to the input audio features at each timestep. The adaptation strategies in \citet{caglayan2019multimodal} match ours more closely: they initialize the encoder and decoder by passing the visual feature vector through a non-linear layer.


While there are several strategies for integrating visual context in multimodal applications, the source of these improvements has so far remained unclear. Recent work in Multimodal Machine Translation \cite{caglayan2019probing} suggests that the visual modality is particularly useful in conditions where the source signal is corrupted and hence insufficient for completing the task. We use these findings as motivation for investigating the usefulness of visual context in multimodal speech recognition. 

\bibliography{example_paper}

\begin{thebibliography}{27}
\providecommand{\natexlab}[1]{#1}
\providecommand{\url}[1]{\texttt{#1}}
\expandafter\ifx\csname urlstyle\endcsname\relax
  \providecommand{\doi}[1]{doi: #1}\else
  \providecommand{\doi}{doi: \begingroup \urlstyle{rm}\Url}\fi

\bibitem[Assael et~al.(2016)Assael, Shillingford, Whiteson, and
  De~Freitas]{assael2016lipnet}
Assael, Y.~M., Shillingford, B., Whiteson, S., and De~Freitas, N.
\newblock Lipnet: End-to-end sentence-level lipreading.
\newblock \emph{arXiv preprint arXiv:1611.01599}, 2016.

\bibitem[Bahdanau et~al.(2014)Bahdanau, Cho, and Bengio]{bahdanau2014neural}
Bahdanau, D., Cho, K., and Bengio, Y.
\newblock Neural machine translation by jointly learning to align and
  translate.
\newblock \emph{arXiv preprint arXiv:1409.0473}, 2014.

\bibitem[Barrault et~al.(2018)Barrault, Bougares, Specia, Lala, Elliott, and
  Frank]{barrault2018findings}
Barrault, L., Bougares, F., Specia, L., Lala, C., Elliott, D., and Frank, S.
\newblock Findings of the third shared task on multimodal machine translation.
\newblock In \emph{THIRD CONFERENCE ON MACHINE TRANSLATION (WMT18)}, volume~2,
  pp.\  308--327, 2018.

\bibitem[Caglayan et~al.(2017)Caglayan, Garc{\'\i}a-Mart{\'\i}nez, Bardet,
  Aransa, Bougares, and Barrault]{caglayan2017nmtpy}
Caglayan, O., Garc{\'\i}a-Mart{\'\i}nez, M., Bardet, A., Aransa, W., Bougares,
  F., and Barrault, L.
\newblock Nmtpy: A flexible toolkit for advanced neural machine translation
  systems.
\newblock \emph{The Prague Bulletin of Mathematical Linguistics}, 109\penalty0
  (1):\penalty0 15--28, 2017.

\bibitem[Caglayan et~al.(2019{\natexlab{a}})Caglayan, Madhyastha, Specia, and
  Barrault]{caglayan2019probing}
Caglayan, O., Madhyastha, P., Specia, L., and Barrault, L.
\newblock Probing the need for visual context in multimodal machine
  translation.
\newblock \emph{arXiv preprint arXiv:1903.08678}, 2019{\natexlab{a}}.

\bibitem[Caglayan et~al.(2019{\natexlab{b}})Caglayan, Sanabria, Palaskar,
  Barraul, and Metze]{caglayan2019multimodal}
Caglayan, O., Sanabria, R., Palaskar, S., Barraul, L., and Metze, F.
\newblock Multimodal grounding for sequence-to-sequence speech recognition.
\newblock In \emph{ICASSP 2019-2019 IEEE International Conference on Acoustics,
  Speech and Signal Processing (ICASSP)}, pp.\  8648--8652. IEEE,
  2019{\natexlab{b}}.

\bibitem[Chan et~al.(2016)Chan, Jaitly, Le, and Vinyals]{chan2016listen}
Chan, W., Jaitly, N., Le, Q., and Vinyals, O.
\newblock Listen, attend and spell: A neural network for large vocabulary
  conversational speech recognition.
\newblock In \emph{2016 IEEE International Conference on Acoustics, Speech and
  Signal Processing (ICASSP)}, pp.\  4960--4964. IEEE, 2016.

\bibitem[Chung et~al.(2017)Chung, Senior, Vinyals, and Zisserman]{chung2017lip}
Chung, J.~S., Senior, A., Vinyals, O., and Zisserman, A.
\newblock Lip reading sentences in the wild.
\newblock In \emph{2017 IEEE Conference on Computer Vision and Pattern
  Recognition (CVPR)}, pp.\  3444--3453. IEEE, 2017.

\bibitem[Elliott(2018)]{elliott2018adversarial}
Elliott, D.
\newblock Adversarial evaluation of multimodal machine translation.
\newblock In \emph{Proceedings of the 2018 Conference on Empirical Methods in
  Natural Language Processing}, pp.\  2974--2978, 2018.

\bibitem[Gupta et~al.(2017)Gupta, Miao, Neves, and Metze]{gupta2017visual}
Gupta, A., Miao, Y., Neves, L., and Metze, F.
\newblock Visual features for context-aware speech recognition.
\newblock In \emph{2017 IEEE International Conference on Acoustics, Speech and
  Signal Processing (ICASSP)}, pp.\  5020--5024. IEEE, 2017.

\bibitem[Harwath \& Glass(2015)Harwath and Glass]{harwath2015deep}
Harwath, D. and Glass, J.
\newblock Deep multimodal semantic embeddings for speech and images.
\newblock In \emph{2015 IEEE Workshop on Automatic Speech Recognition and
  Understanding (ASRU)}, pp.\  237--244. IEEE, 2015.

\bibitem[Harwath et~al.(2016)Harwath, Torralba, and
  Glass]{harwath2016unsupervised}
Harwath, D., Torralba, A., and Glass, J.
\newblock Unsupervised learning of spoken language with visual context.
\newblock In \emph{Advances in Neural Information Processing Systems}, pp.\
  1858--1866, 2016.

\bibitem[He et~al.(2016)He, Zhang, Ren, and Sun]{he2016deep}
He, K., Zhang, X., Ren, S., and Sun, J.
\newblock Deep residual learning for image recognition.
\newblock In \emph{Proceedings of the IEEE conference on computer vision and
  pattern recognition}, pp.\  770--778, 2016.

\bibitem[Kingma \& Ba(2014)Kingma and Ba]{kingma2014adam}
Kingma, D.~P. and Ba, J.
\newblock Adam: A method for stochastic optimization.
\newblock \emph{arXiv preprint arXiv:1412.6980}, 2014.

\bibitem[Libovick{\`y} \& Helcl(2017)Libovick{\`y} and
  Helcl]{libovicky2017attention}
Libovick{\`y}, J. and Helcl, J.
\newblock Attention strategies for multi-source sequence-to-sequence learning.
\newblock \emph{arXiv preprint arXiv:1704.06567}, 2017.

\bibitem[McGurk \& MacDonald(1976)McGurk and MacDonald]{mcgurk1976hearing}
McGurk, H. and MacDonald, J.
\newblock Hearing lips and seeing voices.
\newblock \emph{Nature}, 264\penalty0 (5588):\penalty0 746, 1976.

\bibitem[Miao \& Metze(2016)Miao and Metze]{miao2016open}
Miao, Y. and Metze, F.
\newblock Open-domain audio-visual speech recognition: A deep learning
  approach.
\newblock In \emph{Interspeech}, pp.\  3414--3418, 2016.

\bibitem[Moriya \& Jones(2018)Moriya and Jones]{moriya2018lstm}
Moriya, Y. and Jones, G.~J.
\newblock Lstm language model adaptation with images and titles for multimedia
  automatic speech recognition.
\newblock In \emph{2018 IEEE Spoken Language Technology Workshop (SLT)}, pp.\
  219--226. IEEE, 2018.

\bibitem[Moriya \& Jones(2019)Moriya and Jones]{moriya2019multimodal}
Moriya, Y. and Jones, G.~J.
\newblock Multimodal speaker adaptation of acoustic model and language model
  for asr using speaker face embedding.
\newblock In \emph{ICASSP 2019-2019 IEEE International Conference on Acoustics,
  Speech and Signal Processing (ICASSP)}, pp.\  8643--8647. IEEE, 2019.

\bibitem[Mroueh et~al.(2015)Mroueh, Marcheret, and Goel]{mroueh2015deep}
Mroueh, Y., Marcheret, E., and Goel, V.
\newblock Deep multimodal learning for audio-visual speech recognition.
\newblock In \emph{2015 IEEE International Conference on Acoustics, Speech and
  Signal Processing (ICASSP)}, pp.\  2130--2134. IEEE, 2015.

\bibitem[Palaskar et~al.(2018)Palaskar, Sanabria, and Metze]{palaskar2018end}
Palaskar, S., Sanabria, R., and Metze, F.
\newblock End-to-end multimodal speech recognition.
\newblock In \emph{2018 IEEE International Conference on Acoustics, Speech and
  Signal Processing (ICASSP)}, pp.\  5774--5778. IEEE, 2018.

\bibitem[Pascanu et~al.(2012)Pascanu, Mikolov, and
  Bengio]{pascanu2012understanding}
Pascanu, R., Mikolov, T., and Bengio, Y.
\newblock Understanding the exploding gradient problem.
\newblock \emph{CoRR, abs/1211.5063}, 2, 2012.

\bibitem[Press \& Wolf(2016)Press and Wolf]{press2016using}
Press, O. and Wolf, L.
\newblock Using the output embedding to improve language models.
\newblock \emph{arXiv preprint arXiv:1608.05859}, 2016.

\bibitem[Sanabria et~al.(2018)Sanabria, Caglayan, Palaskar, Elliott, Barrault,
  Specia, and Metze]{sanabria2018how2}
Sanabria, R., Caglayan, O., Palaskar, S., Elliott, D., Barrault, L., Specia,
  L., and Metze, F.
\newblock How2: A large-scale dataset for multimodal language understanding.
\newblock \emph{arXiv preprint arXiv:1811.00347}, 2018.

\bibitem[Sun et~al.(2016)Sun, Harwath, and Glass]{sun2016look}
Sun, F., Harwath, D., and Glass, J.
\newblock Look, listen, and decode: Multimodal speech recognition with images.
\newblock In \emph{2016 IEEE Spoken Language Technology Workshop (SLT)}, pp.\
  573--578. IEEE, 2016.

\bibitem[Toutanova et~al.(2003)Toutanova, Klein, Manning, and
  Singer]{toutanova2003feature}
Toutanova, K., Klein, D., Manning, C.~D., and Singer, Y.
\newblock Feature-rich part-of-speech tagging with a cyclic dependency network.
\newblock In \emph{Proceedings of the 2003 conference of the North American
  chapter of the association for computational linguistics on human language
  technology-volume 1}, pp.\  173--180. Association for computational
  Linguistics, 2003.

\bibitem[Zhou et~al.(2019)Zhou, Yang, Chen, Wang, and Jia]{zhou2019modality}
Zhou, P., Yang, W., Chen, W., Wang, Y., and Jia, J.
\newblock Modality attention for end-to-end audio-visual speech recognition.
\newblock In \emph{ICASSP 2019-2019 IEEE International Conference on Acoustics,
  Speech and Signal Processing (ICASSP)}, pp.\  6565--6569. IEEE, 2019.

\end{thebibliography}
\bibliographystyle{icml2019}

\end{document}